\documentclass{article}

\usepackage[preprint]{decision_pattern}

\usepackage[utf8]{inputenc} 
\usepackage[T1]{fontenc}    
\usepackage{hyperref}       
\usepackage{url}            
\usepackage{booktabs}       
\usepackage{amsfonts}       
\usepackage{amsmath}        
\usepackage{nicefrac}       
\usepackage{microtype}      
\usepackage{xcolor}         
\usepackage{multirow}
\usepackage{multicol}
\usepackage{wrapfig}
\usepackage{float}
\usepackage{graphicx}
\graphicspath{{svg-inkscape/}}

\title{Understanding Generalization through Decision Pattern Shift}

\date{May 12, 2026}

\author{%
  Huiqi Deng \\
  Xi'an Jiaotong University \\
  \And
  Yibo Li \\
  Xi'an Jiaotong University \\
  \And
  Quanshi Zhang \\
  Shanghai Jiao Tong University \\
  \And
  Peng Zhang \\
  Xi'an Jiaotong University \\
  \And
  Hongbin Pei \\
  Xi'an Jiaotong University \\
  \And
  Xia Hu \\
  Shanghai Artificial Intelligence Laboratory \\
}

\begin{document}

\maketitle

\begin{abstract}
Understanding why deep neural networks (DNNs) fail to generalize to unseen samples remains a long-standing challenge. 
Existing studies mainly examine changes in externally observable factors such as data, representations, or outputs, yet offer limited insight into how a model’s internal decision mechanism evolves from training to test. 
To address this gap, we introduce \textbf{Decision Pattern Shift (DPS)}, a new perspective that defines generalization through the stability of internal decision patterns and quantifies failure as their deviation from those learned during training.
Specifically, we represent each sample’s decision pattern as a GradCAM-based channel-contribution vector, which captures how feature channels collectively support a prediction, and we propose the DPS metric to measure its discrepancy from the class-average pattern.
Empirical analyses across multiple datasets and architectures show that, 
(i) decision patterns form a highly structured, class-consistent space with strong intra-class cohesion and low inter-class confusion, enabling direct analysis of a model’s decision logic;
(ii) the DPS magnitude correlates linearly with the generalization gap (nearly all Pearson $r > 0.8$), revealing generalization as a systematic drift in the model’s internal decision mechanism;
(iii) the DPS spectrum organizes diverse generalization degradation scenarios (covering ideal generalization, in-distribution degradation, domain shift, out-of-distribution, and shortcut learning) into a continuous trajectory, providing a unified explanation of their failure modes.
These findings open up new possibilities for early generalization-risk detection, failure-mode diagnosis, and channel-level defect localization.
\end{abstract}    
\section{Introduction}
\label{sec:intro}

Generalization has been studied from multiple perspectives, ranging from theoretical upper bounds based on model capacity~\cite{shalizi2013predictive,bartlett2019nearly,yang2023nearly} to empirical measures of parameter-space geometry and training dynamics~\cite{hochreiter1997flat,dinh2017sharp,fort2020stiffness}, as well as behavioral phenomena such as memorization and shortcut learning~\cite{arpit2017closer,geirhos2020shortcut}.
More recently, mechanism-oriented work has linked generalization failure to external shifts in the data~\cite{zhang2022delving,hendrycks2021many}, representation~\cite{chuang2020estimating,ruan2022optimal}, output behavior~\cite{hochreiter1997flat,novak2018sensitivity}, or the implicit biases induced by model architectures and optimization algorithms
~\cite{l.2018a,keskar2016large,zhang2021interpreting}.

Despite these advances, existing perspectives primarily describe external observable behaviors or factors, leaving the internal decision-making mechanisms that drive generalization largely unexplored.

In this paper, we introduce a new perspective—Decision Pattern Shift (DPS)—for understanding generalization. We argue that \textit{a model’s generalization capacity is fundamentally reflected in its ability to maintain stable and consistent decision patterns on unseen samples, aligned with those formed on the training set.}
When these decision patterns deviate substantially from their training counterparts, the model tends to exhibit generalization failure.

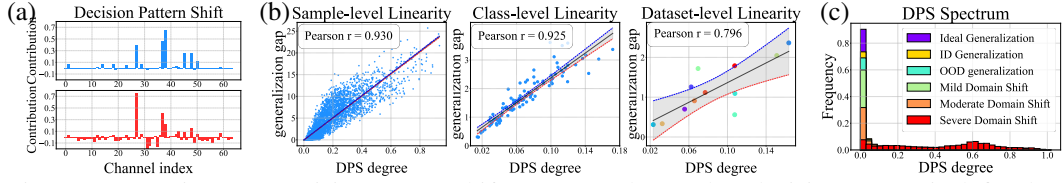
\begin{figure*}[t]
    \centering
  \begingroup
  \def\svgwidth{0.999\textwidth}%
\begingroup%
  \makeatletter%
  \providecommand\color[2][]{%
    \errmessage{(Inkscape) Color is used for the text in Inkscape, but the package 'color.sty' is not loaded}%
    \renewcommand\color[2][]{}%
  }%
  \providecommand\transparent[1]{%
    \errmessage{(Inkscape) Transparency is used (non-zero) for the text in Inkscape, but the package 'transparent.sty' is not loaded}%
    \renewcommand\transparent[1]{}%
  }%
  \providecommand\rotatebox[2]{#2}%
  \newcommand*\fsize{\dimexpr\f@size pt\relax}%
  \newcommand*\lineheight[1]{\fontsize{\fsize}{#1\fsize}\selectfont}%
  \ifx\svgwidth\undefined%
    \setlength{\unitlength}{5064.23327637bp}%
    \ifx\svgscale\undefined%
      \relax%
    \else%
      \setlength{\unitlength}{\unitlength * \real{\svgscale}}%
    \fi%
  \else%
    \setlength{\unitlength}{\svgwidth}%
  \fi%
  \global\let\svgwidth\undefined%
  \global\let\svgscale\undefined%
  \makeatother%
  \begin{picture}(1,0.16349958)%
    \lineheight{1}%
    \setlength\tabcolsep{0pt}%
    \put(0,0){\includegraphics[width=\unitlength,page=1]{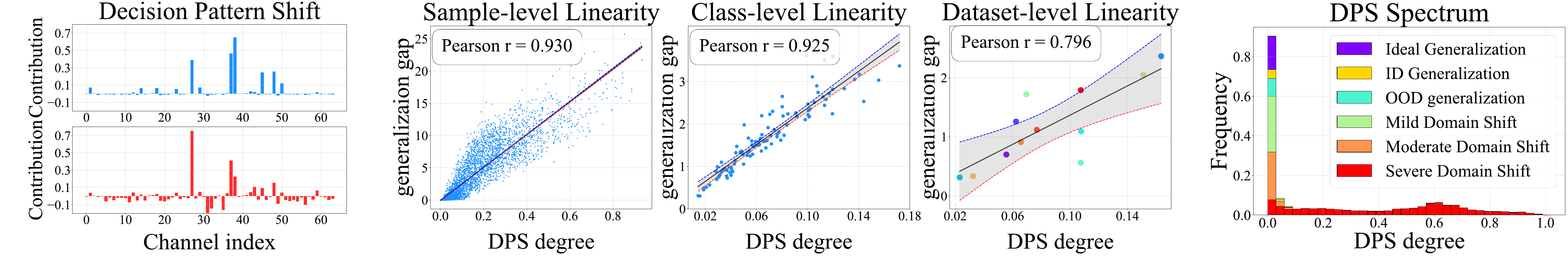}}%
    \put(-0.00050417,0.14928222){\color[rgb]{0,0,0}\makebox(0,0)[lt]{\lineheight{1.25}\smash{\begin{tabular}[t]{l}(a)\end{tabular}}}}%
    \put(0.77028245,0.14928222){\color[rgb]{0,0,0}\makebox(0,0)[lt]{\lineheight{1.25}\smash{\begin{tabular}[t]{l}(c)\end{tabular}}}}%
    \put(0.23871778,0.14928222){\color[rgb]{0,0,0}\makebox(0,0)[lt]{\lineheight{1.25}\smash{\begin{tabular}[t]{l}(b)\end{tabular}}}}%
  \end{picture}%
\endgroup%
  \endgroup

    \vspace{-20pt}
\caption{Overview.
(a) Decision pattern shift (DPS): each sample’s decision pattern is defined as a channel-wise contribution distribution to the model’s prediction.
DPS quantifies the discrepancy between a test sample’s decision pattern (red) and the class-mean decision pattern in the training set, reflecting deviations in the decision pattern space.
(b) Linear relationship: DPS magnitude exhibits a strong linear correlation with the generalization gap across sample-, class-, and dataset levels.
(c) DPS spectrum: different types of generalization failures follow a continuous and coherent evolutionary trajectory on the DPS spectrum. 
}
  \label{fig:overview}
  \vspace{-10pt}
\end{figure*}

\textbf{Characterization of Decision Patterns.}
A key challenge in this perspective is how to faithfully characterize a model's decision pattern for each sample. 
We define the decision pattern as an interpretable, quantitative representation of the model's decision logic. 
To instantiate it, we propose a GradCAM-based formulation that represents each sample's decision pattern as a channel-wise contribution vector (Fig.~\ref{fig:overview}(a)), quantifying how the model distributes reliance across convolutional channels during prediction.

We then verify the faithfulness and structural validity of the proposed decision patterns.
First, the network output is approximately the sum of the decision-pattern vector elements, indicating that this representation reflects the model's internal decision logic.
Second, across architectures and datasets, decision patterns form a structured space with high intra-class cohesion and inter-class separability, with cosine similarities typically above 0.9 within classes and below 0.5 across classes.
These results show that decision patterns provide a faithful and structured representation for quantifying and analyzing internal decision logic.

\textbf{Definition and Measurement of Decision Pattern Shift.}
We define \textit{Decision Pattern Shift (DPS)} as the discrepancy between a model’s decision patterns on the training and test samples.
To quantify this discrepancy, we introduce a DPS metric that measures the cosine dissimilarity between each test sample’s decision pattern and the class-average decision pattern learned from the training set (illustrated in Fig.\ref{fig:overview}(b)).
This formulation can be naturally extended to the class and dataset levels, allowing decision pattern shifts to be analyzed at different granularities.

\textbf{Linear Correlation between DPS and Generalization Gap.}
We further establish the connection between DPS and generalization gap. 
Theoretical analysis demonstrates a strong linear correlation between the two.
This correlation is consistently verified across CIFAR100, TinyImageNet, and ImageNet using representative architectures including VGG, ResNet, and GoogLeNet: 
As shown in Fig.\ref{fig:overview}(b), across all granularities (sample, class, and dataset),  the magnitude of DPS is highly linearly correlated with the generalization gap, with Pearson coefficients typically above 0.8, 
which demonstrates that \textit{generalization gap is strongly associated with systematic drift in the model’s internal decision mechanism from training to test.}

\textbf{DPS Spectrum Reveals the Evolutionary Trajectory of Generalization Degradation.}
Moreover, we discover that the DPS spectrum organizes diverse generalization degradation scenarios (covering ideal generalization, in-distribution degradation, domain shift, out-of-distribution, and shortcut learning) into a continuous trajectory: 
as shown in Fig.~\ref{fig:overview}(c), the DPS distribution gradually shifts rightward, reflecting the continuous degradation of stability in the model’s decision logic.
In contrast to previous studies that treat each failure type separately, the DPS spectrum provides a unified framework for quantitatively describing and comparing various types of generalization failures.

\textbf{Contributions}. 
Our main contributions are as follows: \par
(i) We introduce \textit{Decision Pattern Shift (DPS)}, correlating generalization via internal pattern stability and quantifying failure as pattern deviation between training and unseen samples. \par
(ii) We instantiate a GradCAM-based decision-pattern representation, forming a structured space with strong intra-class cohesion and low inter-class confusion. \par
(iii) We validate, theoretically and empirically, a strong linear correlation between DPS magnitude and generalization gap, establishing DPS as a diagnostic indicator for generalization quality. \par
(iv) We show that various generalization failures follow a continuous trajectory on the DPS spectrum, providing a unified framework for quantifying generalization degradation. \par

\section{Related work}
\textbf{Generalization}. 
Existing studies explain generalization through capacity-based theories (e.g., PAC learning)~\cite{shalizi2013predictive,bartlett2019nearly,yang2023nearly}, properties of parameter space and training dynamics (e.g., loss landscape flatness)~\cite{hochreiter1997flat,li2018visualizing,jastrzebski2021catastrophic,fort2020stiffness}, and behavioral phenomena such as memorization, shortcut learning, and OOD performance~\cite{arpit2017closer,stephenson2021on,geirhos2020shortcut,zhang2012generalization,hendrycks2021many,koh2021wilds}.

Most related to our work is the mechanism-oriented view, which explains generalization through four levels:
(i) the \textit{data level}, emphasizing distribution shift, limited diversity, and spurious correlations~\cite{zhang2022delving,yu2022can,izmailov2022feature};
(ii) the \textit{representation level}, relating failure to unstable, non-invariant, or non-causal features across domains~\cite{ben2006analysis,chuang2020estimating,ruan2022optimal,zhao2019learning,mitrovic2021representation};
(iii) the \textit{output (or loss) level}, linking generalization to prediction or loss stability under perturbations~\cite{dinh2017sharp,novak2018sensitivity}; and
(iv) the \textit{model/optimization level}, examining inductive biases from architectures and optimization strategies~\cite{l.2018a,keskar2016large,deng2022discovering,zhang2021interpreting}.

Unlike these paradigms, which attribute failure to \textit{external observable factors} such as data, representations, output behavior, or optimization dynamics, we focus on whether a model preserves \textit{stable internal decision logic} from training to testing. Specifically, we introduce Decision Pattern Shift (DPS) to quantify changes in decision patterns across datasets.

\textbf{DNN Interpretability.}
Interpretability research aims to understand the internal decision logic of DNNs. Attribution methods such as Shapley Value~\cite{lundberg2017unified} and GradCAM~\cite{selvaraju2017grad} quantify contributions of input features to predictions. Concept-level approaches, including TCAV~\cite{kim2018interpretability}, concept bottleneck models (CBM)~\cite{koh2020concept}, and semantic decision tree~\cite{zhang2019interpreting}, explain reasoning with human-understandable concepts. Network Dissection~\cite{bau2017network} further shows that neurons can encode interpretable semantic entities.

In distinction to the prior works, we explicitly and quantitatively establish a direct connection between interpretability and generalization.

\section{Decision Pattern Characterization}
\label{sec:formatting}
Unlike previous studies that attribute generalization behavior to external factors such as data or feature representation shifts, we examine it from the perspective of the \textit{shift of internal decision logic} in DNNs.
The central challenge lies in defining a faithful representation of a model’s decision logic for individual samples and validating its reliability.

\subsection{Decision Pattern Definition}
\label{sec:Decision Pattern Definition}
We define a decision pattern as a representation that characterizes the model’s internal decision logic for a given sample.
Intuitively, it reflects how the model organizes its internal features to reach a prediction.

Here, we provide a general, formalized representation of this concept.
Formally, for a model $f$ and an input sample $x$, the decision pattern with respect to class $c$ is defined as:
\begin{equation}
    \boldsymbol{p}^c(x)  = \text{DecisionPattern}(f,x,c) 
\end{equation}
where $\text{DecisionPattern}(\cdot)$ denotes an abstract mapping that extracts a description of the model’s decision process.
Its concrete form (e.g. matrix, or graph) depends on the specific instantiation.

\textbf{GradCAM–based instantiation.}
In this work, we instantiate the decision pattern as a \emph{channel contribution vector}.
For a model $f$ and class $c$, we define
\begin{equation}
    p_k^c = w_k^c \cdot \text{GAP}(A^k), \quad
    w_k^c = \frac{1}{Z} \sum_{i,j} \frac{\partial g^c(x)}{\partial A_{ij}^k},
\end{equation}
where $A^k$ is the activation map of the channel $k$, $\text{GAP}(\cdot)$ denotes global average pooling, $Z=H\times W$ is the number of spatial locations in $A^k$, and $g^c(x)$ is the class-$c$ logit.

The scalar $w_k^c$ quantifies how sensitive the class-$c$ output is to perturbations in channel $k$, 
a property that has been well established in prior works~\cite{selvaraju2017grad}.
Correspondingly, $p_k^c$ captures the overall contribution of that channel after aggregating its spatial activations.
Thus, the derived decision pattern vector $\boldsymbol{p}^c(x) = [p_1^c, \ldots, p_K^c]^\top \in \mathbb{R}^K$ provides a compact, quantitative expression of how the model distributes its reliance across channels, revealing its internal feature-dependency structure.

\begin{wrapfigure}{r}{0.58\columnwidth}
    \vspace{-10pt}
    \centering
  \begingroup
  \def\svgwidth{0.9\linewidth}%
  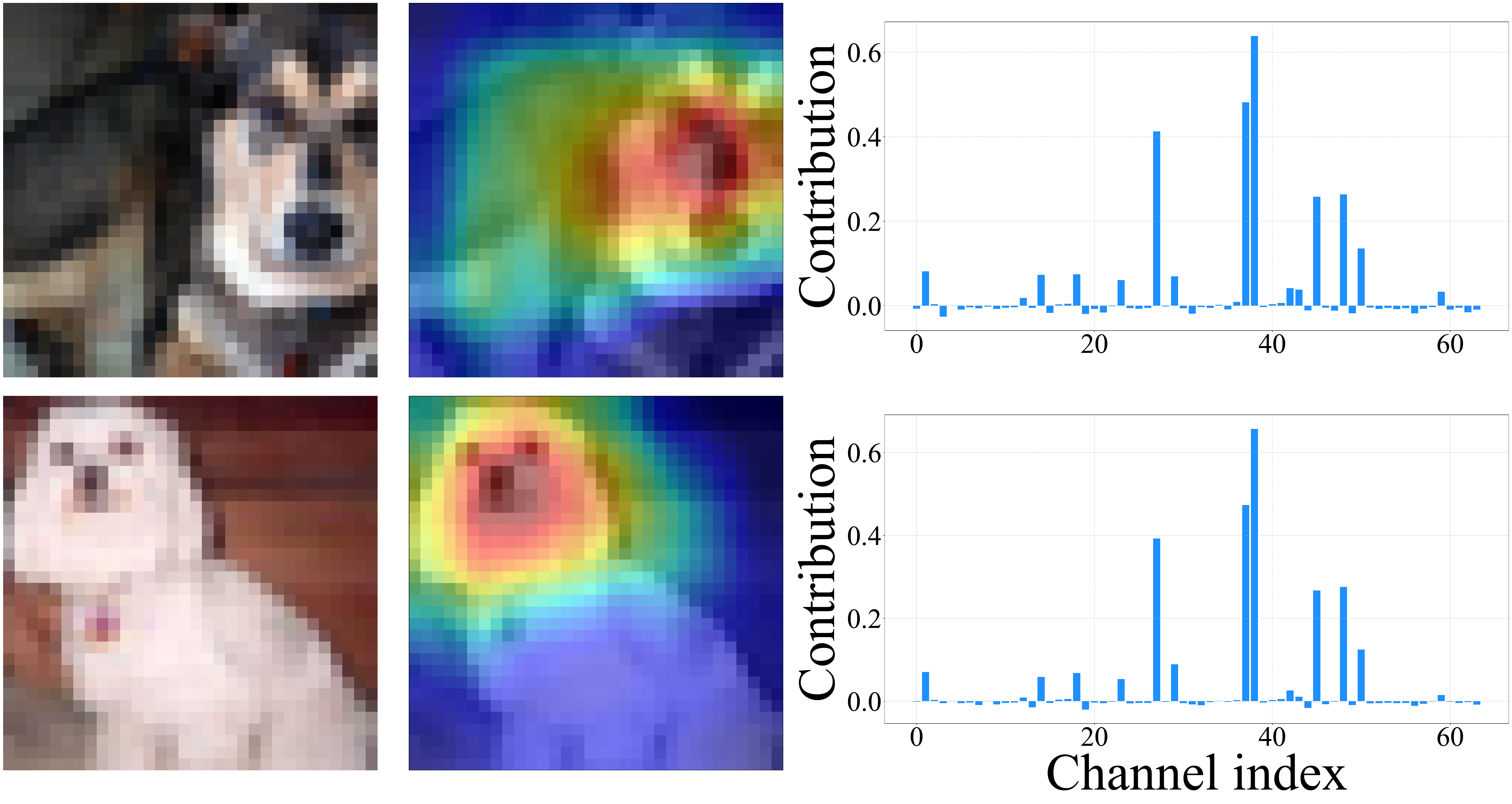%
  \endgroup

    \vspace{-8pt}
    \caption{Decision pattern (right) captures channel-wise contributions, whereas Grad-CAM (middle) highlights salient input regions.}
    \label{fig:DP_illustration}
    \vspace{-12pt}
\end{wrapfigure}

Fig.\ref{fig:DP_illustration} illustrates the decision pattern of a ResNet18 model on two CIFAR10   “dog” samples.
Only a small subset of channels exhibits strong values, while most remain weak, indicating a sparse yet structured decision-dependency pattern.
Moreover, visually different samples share a highly similar decision pattern, implying that the model forms a stable, class-specific dependency structure, an essential property that underlies its generalization behavior.

\textbf{Difference from GradCAM.} 
Our instantiation is derived from the core formulation of GradCAM~\cite{selvaraju2017grad}.
By rearranging the GradCAM expression, we obtain:
\begin{equation}
    a_{ij}^c = ReLU \big(\sum\nolimits_k w_k^c A^k_{ij} \big), 
\end{equation}
Despite this mathematical connection, our formulation differs fundamentally in both semantics and faithfulness.

\noindent (i) Semantic:
Comparing the two formulations, the GradCAM map aggregates channel responses \emph{spatially} to highlight salient input regions, thereby capturing \emph{where the model looks}.
In contrast, the decision pattern aggregates spatial activations across positions to quantify internal feature dependencies at the channel level, thereby capturing \emph{what the model relies on}.
Fig.~\ref{fig:DP_illustration} illustrates this difference, showing that the two representations offer distinct perspectives on the model’s decision behavior.

\noindent (ii) Faithfulness to decision logic:
GradCAM does not directly characterize how internal features jointly support a prediction.
Empirically, two same-class samples with highly similar decision patterns may display substantially different GradCAM maps (Fig.~\ref{fig:DP_illustration}), indicating that spatial saliency alone is not a reliable indicator of whether the model employs similar internal reasoning.

Therefore, while inspired by GradCAM, our representation offers a more faithful, structurally coherent characterization of the model’s decision logic.
Note that our framework is general, allowing the decision pattern to be instantiated using other similar attribution methods as well.

\subsection{Reliability of Decision Pattern Definition}
\label{sec:Reliability of Decision Pattern Definition}
To ensure the decision pattern faithfully represents the model’s internal decision logic, we evaluate it from three complementary perspectives:
(i) Faithfulness: assessing whether it faithfully reflects the predicted output of the model;
(ii) Intra-class consistency: examining whether same-class samples exhibit similar patterns; and
(iii) Inter-class separability: evaluating whether patterns across different classes remain distinguishable.

\textbf{Faithfulness to model's prediction}
Under a $\text{GAP}(\cdot)$ followed by a linear classification head, the class logit admits a channel-wise decomposition up to the bias term and a constant scaling factor. 
More generally, for architectures with nonlinear heads, the proposed decision-pattern vector can be interpreted as a first-order local approximation of the channel-wise evidence supporting the class logit. 
Therefore, the sum of its components provides an evidence-aligned proxy for the model’s prediction, rather than an exact reconstruction of the full decision process.
\begin{equation}
    \label{eqn:approximation} 
         f^c(x) \approx \sum\nolimits_k w_k^c \cdot GAP(A^k) = \sum\nolimits_k p_k^c
    \end{equation}

\textbf{Intra-class consistency and inter-class separability}.
We conduct experiments to validate whether the proposed decision pattern exhibits high intra-class consistency and inter-class separability. To quantify these properties, we first define two evaluation metrics. 

The \textit{intra-class consistency} metric measures the similarity of decision patterns within the same class:
\begin{equation}
    S_{intra}(c) = \frac{1}{\binom{n_c}{2}}\sum\nolimits_{i\neq j,x_i,x_j \in \mathcal{N}_c} cos<\boldsymbol{p}^{c}(x_i), \boldsymbol{p}^{c}(x_j)>
\end{equation}
where $\mathcal{N}_c$ denotes the set of samples belonging to class $c$, and $cos<\cdot, \cdot>$ computes the cosine similarity between two decision pattern vectors. 
A high $S_{intra}(c)$ indicates stronger intra-class consistency, suggesting that the model forms stable and shared  decision structures for that class.

The \textit{inter-class confusability} metric for class $c$ quantifies its maximum similarity to other classes in the decision-pattern space:
\begin{equation}
    S_{inter}(c) = \max_{c' \neq c} cos<\boldsymbol{\overline p}^c, \boldsymbol{\overline p}^{c'}>
\end{equation}
where $\boldsymbol{\overline p}^c = \mathbb{E}_{x_i\in \mathcal{N}_c} [\boldsymbol{p}^{c}(x_i)]$ represents the average decision pattern of class $c$. 
A lower $S_{inter}(c)$ indicates better separability and more distinct decision logic between classes.

\begin{table*}[t]
\vspace{-6pt}
\centering
\caption{Intra-class consistency and inter-class confusability  on ResNet: decision patterns outperform activation patterns across all datasets. The results on VGG and GoogLeNet are in the appendix.}
\label{tab:intra_inter}
\resizebox{\textwidth}{!}{
\begin{tabular}{c|cccc|cccc}
\toprule
\multirow{2}{*}{\textbf{Model}} &
\multicolumn{4}{c|}{Intra-class consistency $\uparrow$} &
\multicolumn{4}{c}{Inter-class confusability  $\downarrow$} \\
\cline{2-9}
& CIFAR-10 & CIFAR-100 & TinyImageNet & ImageNet & CIFAR-10 & CIFAR-100 & TinyImageNet & ImageNet \\
\midrule
Activation Pattern    & 0.894 & 0.810  & 0.776 & 0.875 & 0.542 & 0.710 & 0.450 & 0.551 \\
Decision Pattern       & \textbf{0.973} & \textbf{0.925} & \textbf{0.938} & \textbf{0.960} & \textbf{0.109} & \textbf{0.548} & \textbf{0.323} & \textbf{0.307} \\
\bottomrule
\end{tabular}
}
\vspace{-10pt}
\end{table*}

We conduct these evaluations on four datasets, namely CIFAR10, CIFAR100~\cite{krizhevsky2009learning}, TinyImageNet~\cite{Le2015TinyIV}, and ImageNet~\cite{deng2009imagenet}, using architectures including VGG~\cite{simonyan2014very}, ResNet~\cite{he2016deep}, and GoogLeNet~\cite{szegedy2015going}.
Across all datasets and architectures, we consistently observe the following phenomena:

\noindent (i) \textbf{High intra-class consistency.}
Decision patterns exhibit high intra-class consistency, with average cosine similarity exceeding 0.9 (Tab.~\ref{tab:intra_inter}). 
This indicates that despite varying appearances, same-class samples evoke nearly identical internal decision patterns for inference.

\noindent (ii) \textbf{Low inter-class confusion.}
Patterns across different classes are well-separated, with average inter-class confusability below 0.5 (Tab.~\ref{tab:intra_inter}), implying distinct decision boundaries across categories.


\begin{wrapfigure}{r}{0.58\columnwidth}
    \vspace{-4pt}
    \centering
  \begingroup
  \def\svgwidth{0.9\linewidth}%
  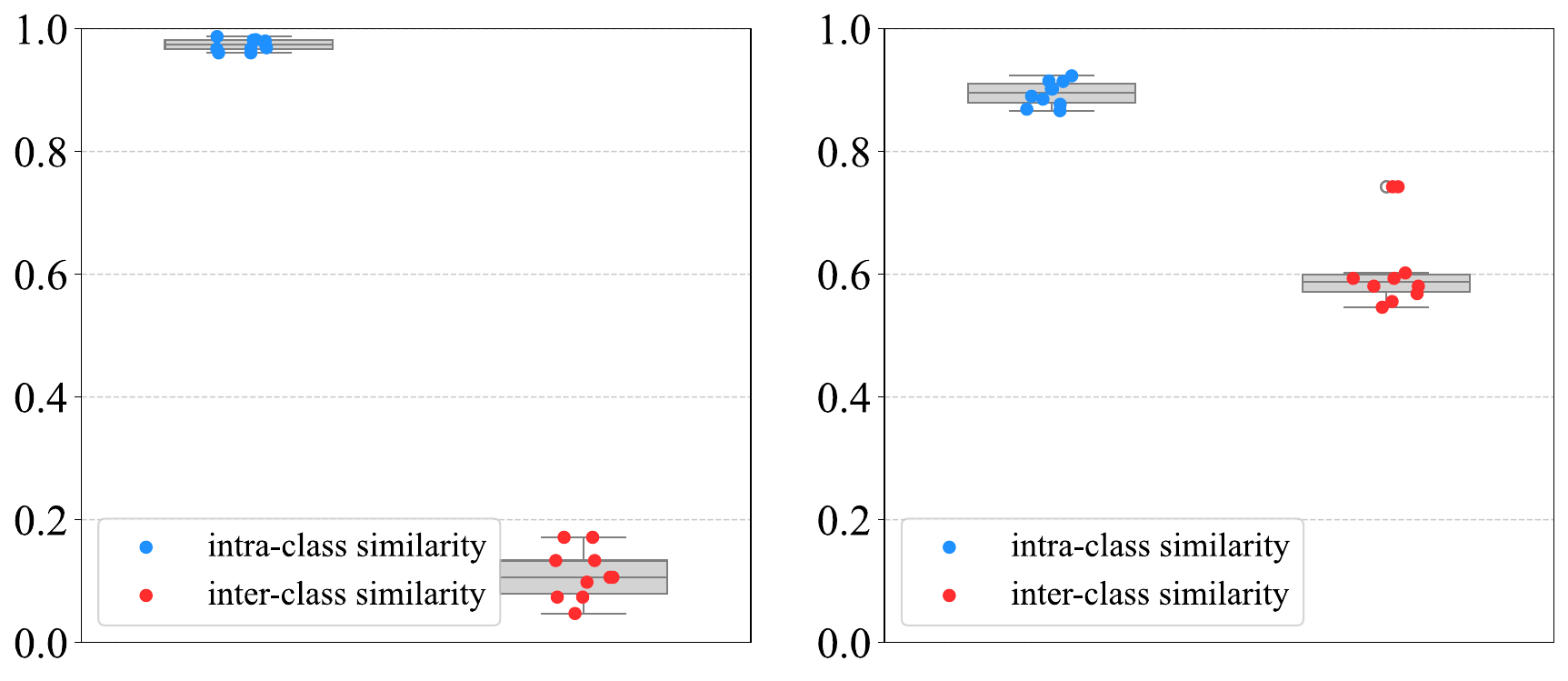%
  \endgroup

     \caption{Comparison between decision patterns (left) and the activation-pattern baseline (right). Box-plots of intra-class similarity and inter-class similarity. Each point corresponds to a single class.}
  \label{fig:intra-class consistency and inter-class separability 0}
  \vspace{-10pt}
\end{wrapfigure}

\noindent (iii) \textbf{Advantage over activation patterns.}
To differentiate the decision-pattern space from feature representation space, we compare against an activation-based baseline $\boldsymbol{a} = [GAP(A^1), \dots, GAP(A^K)]$. 
Capturing class-agnostic features rather than class-specific evidence, these activations exhibit lower coherence and weaker separation, resulting in an entangled representation space(Tab.~\ref{tab:intra_inter}).

\begin{figure}[t]
    \vspace{-12pt}
    \centering
    \includegraphics[width=\textwidth]{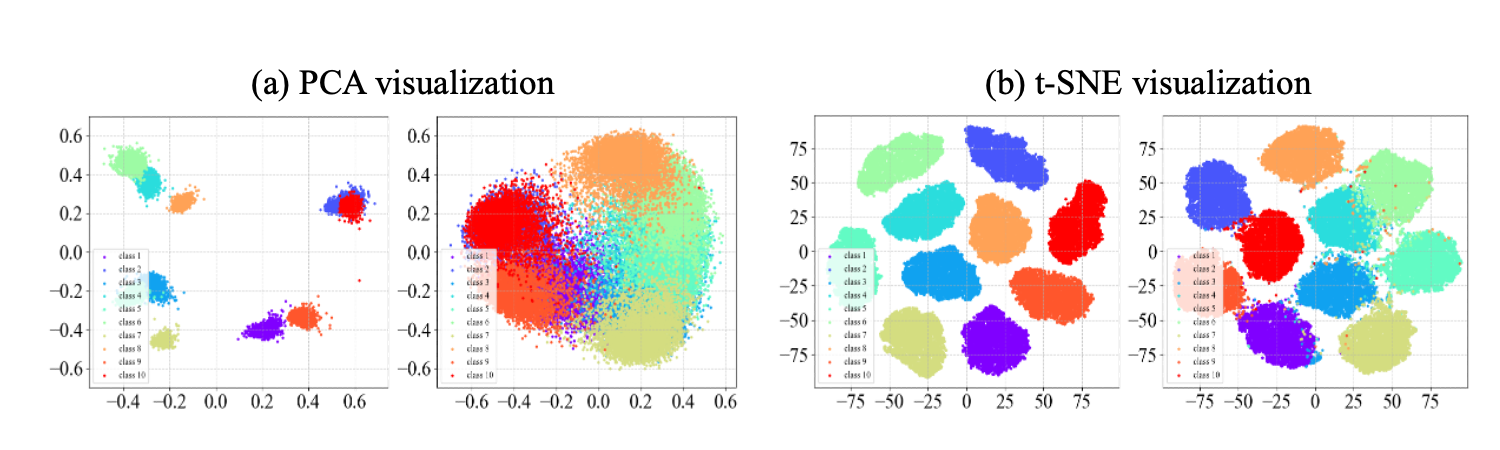}
    \vspace{-18pt}
     \caption{
Visualization of decision patterns (left) and the activation-pattern baseline (right) through PCA and t-SNE.
(a) PCA projections of all samples’ decision pattern vectors, with each color indicating a class.
(b) t-SNE projections of decision pattern vectors.
}
  \label{fig:intra-class consistency and inter-class separability 1}
  \vspace{-10pt}
\end{figure}

For illustration, Fig.~\ref{fig:intra-class consistency and inter-class separability 0} and Fig.~\ref{fig:intra-class consistency and inter-class separability 1} shows representative results on CIFAR10 with ResNet20, which clearly exhibit high intra-class consistency, low inter-class confusion, structured clustering, and cleaner structure than activation patterns. 
Based on the above results, we claim that the \textit{decision-pattern space} can capture how a model organizes evidence for prediction and provides a more diagnostic lens on generalization. In this space, same-class samples may differ in appearance yet invoke similar decision pathways, while divergence of decision patterns on unseen data signals generalization deterioration. 
Comprehensive results for other datasets and models are provided in the appendix.

\section{Understanding Generalization through Decision Pattern Shift}

As discussed in Sec.~\ref{sec:Reliability of Decision Pattern Definition}, the decision-pattern space offers a more intrinsic and mechanism-grounded perspective for generalization.
Building on this foundation, we connect generalization with \textbf{\textit{Decision Pattern Shift (DPS)}}, which characterizes how the model’s internal decision logic evolves from training to unseen data.

\subsection{Defining and Measuring Decision Pattern Shift}
We define Decision Pattern Shift (DPS) as the systematic deviation of decision patterns (channel-wise contribution vectors) between training and test sets. 
To quantify this deviation, we introduce hierarchical metrics evaluating DPS at sample, class, and dataset levels.

To quantify the deviation for each test sample, we take the class-wise mean decision pattern from the training set as the reference, given the strong intra-class consistency observed in decision patterns.
For a test sample $x$ belonging to class $y$, the sample-level shift is defined as:
\begin{equation}
\label{eqn:def_DPS}
\text{DPS}_{\text{sample}}(x) = 1 - \cos<\boldsymbol{p}^c(x), \boldsymbol{\overline{p}}^{c}>,
\end{equation}
where $\boldsymbol{p}^c(x)$ denotes the decision pattern of the test sample $x$ with respect to its ground truth label $c$, and $\boldsymbol{\overline p}^c = \mathbb{E}_{x' \in \mathcal{D}_{train}^c} [\boldsymbol{p}^c(x')]$  is the average pattern of class $c$ computed from the training set.

This sample-level measure can be naturally extended to characterize aggregated deviations across samples and classes.
The class-level DPS is defined as the mean shift over all test samples of class $c$:
\begin{equation}
\label{eqn:def_clsDPS}
\text{DPS}_{\text{class}}(c) = \frac{1}{|\mathcal{D}^c_{test}|}\sum\nolimits_{x \in \mathcal{D}^c_{test}} \text{DPS}_{\text{sample}}(x),
\end{equation}
and the dataset-level DPS quantifies the global drift between training and test distributions:
\begin{equation}
\label{eqn:def_datasetDPS}
\text{DPS}_{\text{dataset}} = \frac{1}{|\mathcal{D}^{test}|}\sum\nolimits_{x \in \mathcal{D}^{test}} \text{DPS}_{\text{sample}}(x).
\end{equation}

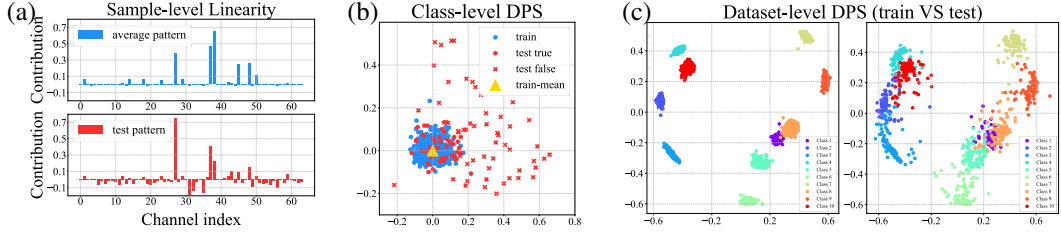
\begin{figure*}[t]
  \centering
  \begingroup
  \def\svgwidth{0.999\textwidth}%
\begingroup%
  \makeatletter%
  \providecommand\color[2][]{%
    \errmessage{(Inkscape) Color is used for the text in Inkscape, but the package 'color.sty' is not loaded}%
    \renewcommand\color[2][]{}%
  }%
  \providecommand\transparent[1]{%
    \errmessage{(Inkscape) Transparency is used (non-zero) for the text in Inkscape, but the package 'transparent.sty' is not loaded}%
    \renewcommand\transparent[1]{}%
  }%
  \providecommand\rotatebox[2]{#2}%
  \newcommand*\fsize{\dimexpr\f@size pt\relax}%
  \newcommand*\lineheight[1]{\fontsize{\fsize}{#1\fsize}\selectfont}%
  \ifx\svgwidth\undefined%
    \setlength{\unitlength}{4485.59875488bp}%
    \ifx\svgscale\undefined%
      \relax%
    \else%
      \setlength{\unitlength}{\unitlength * \real{\svgscale}}%
    \fi%
  \else%
    \setlength{\unitlength}{\svgwidth}%
  \fi%
  \global\let\svgwidth\undefined%
  \global\let\svgscale\undefined%
  \makeatother%
  \begin{picture}(1,0.21719508)%
    \lineheight{1}%
    \setlength\tabcolsep{0pt}%
    \put(0,0){\includegraphics[width=\unitlength,page=1]{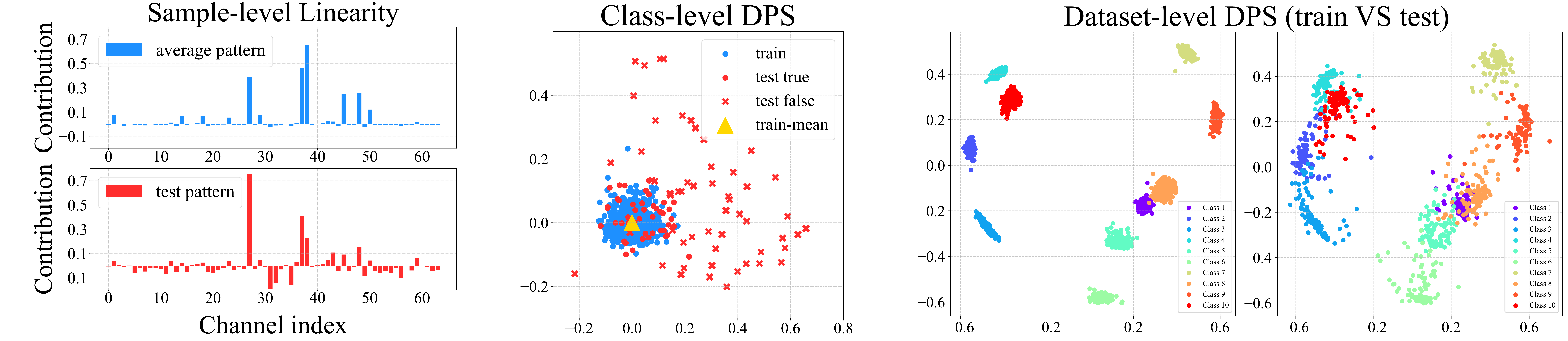}}%
    \put(-0.00063093,0.1996389){\color[rgb]{0,0,0}\makebox(0,0)[lt]{\lineheight{1.25}\smash{\begin{tabular}[t]{l}(a)\end{tabular}}}}%
    \put(0.32240488,0.1996389){\color[rgb]{0,0,0}\makebox(0,0)[lt]{\lineheight{1.25}\smash{\begin{tabular}[t]{l}(b)\end{tabular}}}}%
    \put(0.58267448,0.1996389){\color[rgb]{0,0,0}\makebox(0,0)[lt]{\lineheight{1.25}\smash{\begin{tabular}[t]{l}(c)\end{tabular}}}}%
  \end{picture}%
\endgroup%
  \endgroup

  \vspace{-15pt}
  \caption{Illustration of Decision Pattern Shift (DPS) across multiple levels.
(a) Sample-level DPS: the decision pattern of an individual test sample (red) compared with the class-average training pattern (blue), capturing fine-grained deviations at the sample level.
(b) Class-level DPS: aggregating sample-level DPS reveals class-level structure, i.e., correct test samples cluster tightly around their class mean, while misclassified samples tend to appear near or beyond class boundaries.
(c) Dataset-level DPS: integrating class-level shifts further exposes coarse-grained dataset structure, showing how accumulated deviations across classes shape the global confusion phenonmenon.
}
\label{fig:DPS_illustration}
\vspace{-15pt}
\end{figure*}

To illustrate DPS intuitively, Fig.~\ref{fig:DPS_illustration}(a) presents a sample-level case: 
comparing a misclassified test sample's pattern with the training class-mean reveals a clear structural deviation. 
Despite shared class membership, substantial pattern differences reveal shifted internal reasoning, leading to misclassification. 
Broadly, Fig.~\ref{fig:DPS_illustration}(b)-(c) show sample-level shifts reshaping class- and dataset-level structures: correctly classified samples align with training clusters, while misclassified ones (with larger DPS) drift toward or beyond boundaries, forming global confusion patterns.

\subsection{Linear Correlation between DPS and Generalization Gap}
\label{sec:proof}
We next establish the connection between \textbf{DPS} and the model’s generalization gap.

\textbf{Theoretical Analysis of Linear Correlation.}
The \textit{per-sample generalization gap} is the difference between the empirical loss of a test sample $x$ and its class's average training loss:
\begin{equation}
\label{eqn:def_gengap}
\Delta \mathcal{L}(x) = \mathcal{L}(x) - \mathbb{E}_{x' \in \mathcal{D}^c_{\text{train}}} [\mathcal{L}(x')],
\end{equation}

Since the average training loss is a constant, we analyze the cross-entropy loss $\mathcal{L}(x) = -\log P_c$, where $P_c = \sigma(g_c)$ denotes the softmax probability of the true class logit $g_c$.
Although $\mathcal{L}$ depends on all logits via softmax normalization, its variation is most sensitive along the $g_c$ coordinate.
Accordingly, we perform a local \emph{coordinate-wise} analysis along $g_c$ while holding $\{g_k\}_{k\neq c}$ fixed.
The first-order and second-order  derivatives with respect to $g_c$ are:
\begin{equation}
    \frac{\partial \mathcal{L}}{\partial g_c} = P_c - 1, \quad \frac{\partial^2\mathcal{L}}{\partial g_c^2} = P_c(1-P_c).
\end{equation}

Let $\overline g_c = \mathbb{E}_{x' \in \mathcal{D}^c_{\text{train}}} [g_c]$ denote 
class-wise mean logit in the training set. 
Expanding $\mathcal{L}(x)$ at the reference $\overline g_c$ gives: 
\begin{equation}
\begin{aligned}
   \mathcal{L}(x) & \approx -log \overline P_c + (\overline P_c - 1) \cdot (g_c - \overline g_c) + \frac{1}{2}  \overline P_c (1-\overline P_c)\cdot (g_c - \overline g_c)^2    
\end{aligned}
\end{equation}
As mentioned in Eq.~(\ref{eqn:approximation}), each logit can be decomposed into channel-wise contributions:
\begin{equation}
     g_c(x) \approx  \sum\nolimits_k p_k^c(x), \quad \overline g_c \approx  \sum\nolimits_k \overline p_k^c,
\end{equation}
hence $g_c - \overline g_c \approx  \sum\nolimits_k  (p_k^c(x) - \overline p_k^c)$.
Substituting into the expansion yields:
\begin{equation}
\begin{aligned}
\label{eqn:expansion}
\!\!\! & \mathcal{L}(x)  \approx -log \overline P_c + (\overline P_c - 1) \sum\nolimits_k (p_k^c(x) - \overline p_k^c) + \frac{1}{2}  \overline P_c (1-\overline P_c)\cdot \big(\sum\nolimits_k [p_k^c(x) - \overline p_k^c]\big)^2
\end{aligned}
\end{equation}

\textit{Implication}. 
Eq.~(\ref{eqn:expansion}) links the per-sample generalization gap $\Delta \mathcal{L}(x)$ to the aggregated decision pattern deviation, $\sum_k (p_k^c(x) - \overline p_k^c)$. 
Linearity is governed by the quadratic remainder; since $\tfrac12\,\overline P_c(1-\overline P_c)\le 0.125$, the second-order term is bounded by $0.125 (g_c-\overline g_c)^2$. 
For modest $|g_c-\overline g_c|$, the first-order term dominates, yielding an approximately linear dependence on pattern deviation.

The aggregated deviation above measures how far the current pattern $\boldsymbol{p}^c(x)$ departs from the average training pattern $\boldsymbol{\overline p}^c$, 
whereas our cosine-based DPS in Eq.~(\ref{eqn:def_DPS}) captures the corresponding angular discrepancy in a normalized and scale-invariant manner. 
Although these two quantities are not identical, they are typically monotonically aligned under normalized pattern representations and exhibit an approximately linear relationship in the empirical regimes studied here. 
This provides a mechanistic explanation for the observed linear correlation between DPS(x) and $\Delta \mathcal{L}(x)$, which is further verified in our experiments.


\begin{figure*}[t]
  \centering
  \begingroup
  \def\svgwidth{0.999\textwidth}%
  \input{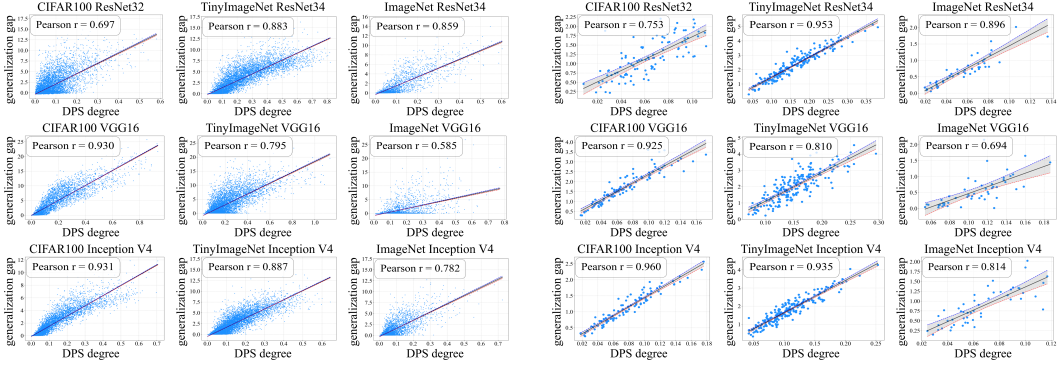}%
  \endgroup

  \vspace{-15pt}
     \caption{Linear correlation between Decision Pattern Shift (DPS) and the generalization gap on CIFAR100, TinyImageNet, and ImageNet across multiple architectures, including ResNet-32/34, VGG16, and GoogLeNet (InceptionV4).
(a) Sample-level correlation between per-sample DPS and generalization gap. Each point corresponds to a test sample, and the fitted line shows the overall linear trend.
(b) Class-level correlation between the average DPS and the class-level generalization gap. Each point represents a single class in the dataset.}
  \label{fig:linear}
  \vspace{-7pt}
\end{figure*}

\textbf{Experimental Validation of Linear Correlation.}
To empirically verify the above theoretical derivation, we examine the linear dependence between $\text{DPS}(x)$ and $\Delta\mathcal{L}(x)$ across datasets (CIFAR100, TinyImageNet, ImageNet) and architectures (VGG, ResNet, GoogLeNet). For ImageNet, we train models on a 50-class subset for computational feasibility.
For comparison, we also evaluate the linear correlation of the activation-based baseline, with results deferred to the appendix.

\noindent \textit{(i) Sample-level validation.} 
We compute DPS and per-sample generalization gaps via Eqs.~(\ref{eqn:def_DPS}) and (\ref{eqn:def_gengap}). 
As shown in Fig.~\ref{fig:linear}(a), linear regression reveals strong correlations, with Pearson coefficients $r \ge 0.78$ ($p<0.001$) across datasets and architectures. 
Even on ImageNet, the correlation remains substantial, though slightly lower due to greater inter-class confusion.

\noindent \textit{(ii) Class-level and dataset-level validation.}
We also compute the \textbf{class-level generalization gap} (mean test-train loss difference) and \textbf{dataset-level generalization error} (gap across all samples), with corresponding DPS values from Eqs.~(\ref{eqn:def_clsDPS}) and (\ref{eqn:def_datasetDPS}). 
As shown in Figs.~\ref{fig:linear}(b) and \ref{fig:DPS_spectrum}(a), a strong linear trend persists at these granularities, indicating that DPS consistently correlates with generalization variations from individual samples to entire datasets.

\textbf{Summary.}
Theoretical and empirical results demonstrate a strong linear correlation between DPS and the generalization gap.
\textbf{This reframes generalization as decision-pattern consistency}: models generalize well when internal logic remains stable across training and unseen samples, failing as consistency deteriorates. 
This unified lens facilitates analyzing generalization across granularities and failure modes, as explored in the next section.
\section{DPS Spectrum Reveals the Evolutionary Trajectory of Generalization Degradation}

Beyond the linear correspondence between DPS and generalization gap, these shifts form a structured distribution unifying diverse manifestations of generalization degradation. 
We define this as \textbf{Decision Pattern Shift Spectrum (DPS Spectrum)}, characterizing the distribution of DPS magnitudes across test samples to reveal how deviations occur collectively or selectively as generalization degrades.

Under this formulation, an ideal model exhibits a DPS spectrum concentrated near zero. To study how the spectrum evolves with degradation, we use VGG16 to construct five representative scenarios and visualize their test-sample decision-pattern shifts. Experimental setups are as follows:
(i) \textbf{Ideal generalization}: Train and test on CIFAR10 classes with strong performance (accuracy gap $<$ 3\%), indicating stable decision logic.
(ii) \textbf{In-distribution generalization}: Train and test on CIFAR10 classes with larger accuracy gaps ($>$ 10\%), representing mild in-distribution degradation.
(iii) \textbf{Domain shift}: Train on CIFAR10 and test on CIFAR10-C at highest severity. We use three corruption types—\textit{snow} (mild, $\sim$25\% drop), \textit{gaussian\_blur} (moderate, $\sim$50\% drop), and \textit{contrast} (severe, $\sim$70\% drop)—to form a continuous evolution of progressive degradation.
(iv) \textbf{OOD generalization}: Train on CIFAR10 and test on resized STL-10 (32×32), representing an OOD setting with limited degradation ($<$ 20\%).
(v) \textbf{Shortcut learning}: Train on Colored MNIST with strong color-label correlation and test on decorrelated data, capturing degradation from reliance on spurious features.

\begin{figure*}[t]
  \centering
  \begingroup
  \def\svgwidth{0.95\textwidth}%
\begingroup%
  \makeatletter%
  \providecommand\color[2][]{%
    \errmessage{(Inkscape) Color is used for the text in Inkscape, but the package 'color.sty' is not loaded}%
    \renewcommand\color[2][]{}%
  }%
  \providecommand\transparent[1]{%
    \errmessage{(Inkscape) Transparency is used (non-zero) for the text in Inkscape, but the package 'transparent.sty' is not loaded}%
    \renewcommand\transparent[1]{}%
  }%
  \providecommand\rotatebox[2]{#2}%
  \newcommand*\fsize{\dimexpr\f@size pt\relax}%
  \newcommand*\lineheight[1]{\fontsize{\fsize}{#1\fsize}\selectfont}%
  \ifx\svgwidth\undefined%
    \setlength{\unitlength}{3864.22265625bp}%
    \ifx\svgscale\undefined%
      \relax%
    \else%
      \setlength{\unitlength}{\unitlength * \real{\svgscale}}%
    \fi%
  \else%
    \setlength{\unitlength}{\svgwidth}%
  \fi%
  \global\let\svgwidth\undefined%
  \global\let\svgscale\undefined%
  \makeatother%
  \begin{picture}(1,0.20165763)%
    \lineheight{1}%
    \setlength\tabcolsep{0pt}%
    \put(0,0){\includegraphics[width=\unitlength,page=1]{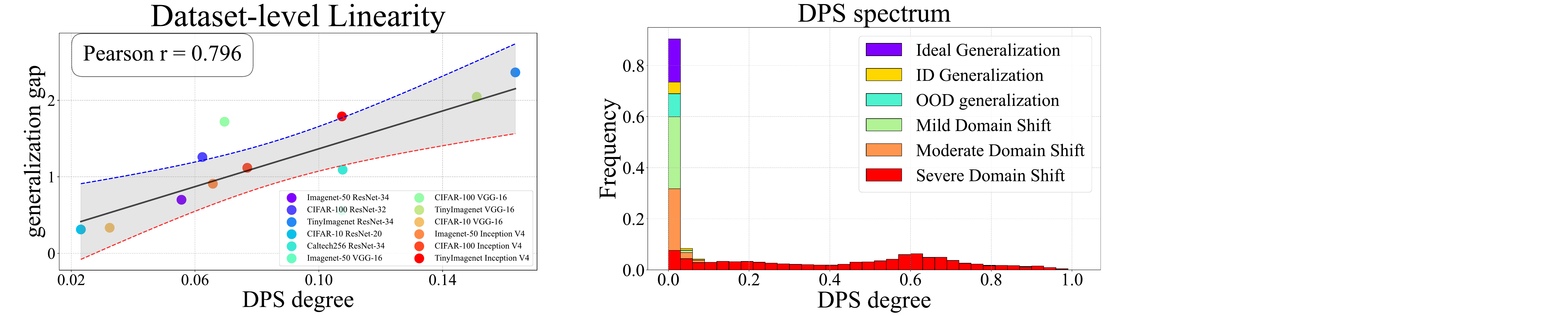}}%
    \put(-0.00066073,0.18302517){\color[rgb]{0,0,0}\makebox(0,0)[lt]{\lineheight{1.25}\smash{\begin{tabular}[t]{l}(a)\end{tabular}}}}%
    \put(0.37816449,0.18302517){\color[rgb]{0,0,0}\makebox(0,0)[lt]{\lineheight{1.25}\smash{\begin{tabular}[t]{l}(b)\end{tabular}}}}%
    \put(0,0){\includegraphics[width=\unitlength,page=2]{DPS_spectrum_1114_svg-tex.pdf}}%
  \end{picture}%
\endgroup%
  \endgroup

  \vspace{-10pt}
     \caption{(a) Dataset-level correlation between the average DPS and the overall generalization gap. Each point corresponds to a dataset–architecture pair. 
      (b) DPS spectra across representative regimes, including ideal, in-distribution, OOD, and mild to severe domain shifts. Spectra exhibit continuous evolution, shifting rightward and contracting as generalization degrades. 
      In shortcut learning, besides the near-zero peak, the spectrum separates into distinct modes, revealing heterogeneous patterns induced by color shortcuts.
     }
  \label{fig:DPS_spectrum}
  \vspace{-6pt}
\end{figure*}

As illustrated in Fig.~\ref{fig:DPS_spectrum}, the DPS spectra under different generalization scenarios reveal several key observations: 

\noindent\textbf{(i) Continuous degradation spectrum with a stable OOD exception:} 
As shown in Fig.~\ref{fig:DPS_spectrum}(b), the DPS spectrum evolves continuously across generalization regimes: under ideal generalization, it is concentrated near zero, while from in-distribution to increasingly severe domain shift, the spectral mass moves rightward and the near-zero peak diminishes, indicating growing instability and an increasing fraction of shifted samples. 
In contrast, for semantically similar OOD data (e.g., STL-10), the spectrum remains close to the in-distribution case without noticeable rightward shift, suggesting that OOD conditions do not necessarily induce decision-pattern shift or generalization failure.

\noindent\textbf{(ii) Structural collapse in shortcut learning:}
The DPS spectrum under shortcut learning exhibits multiple peaks. 
This occurs because the model associates each color with a distinct decision pattern during training; when tested on randomly colored digits, different colors activate different pathways, producing multiple clusters of decision-pattern shifts.
This case represents an extreme form of domain shift, where disrupted feature-label correlations collapse coherent decision logic.

Taken together, these results show that generalization degradation can be viewed as a progressive evolution of the \textbf{DPS Spectrum}, 
from coherent alignment, to systematic drift, and finally to structural collapse. 
This spectral evolution unifies diverse manifestations of generalization failure and provides new insight into the internal dynamics underlying generalization degradation.

\section{Failures and Limitations}
\label{sec:Limitations}

As defined in Eq.~\ref{eqn:def_DPS}, DPS relies on the class-average decision pattern computed from clean training samples as the reference for measuring decision shifts. 
When training labels are corrupted, the class-conditional decision patterns may become inconsistent or multi-modal, thereby reducing the diagnostic power of DPS. 
In such scenarios, deviations from the reference pattern do not necessarily indicate generalization failure, and the DPS–generalization correlation is expected to weaken. 
Empirically, on CIFAR100 with VGG16, 20\% label noise reduces the sample-level correlation from 0.930 to 0.855, consistent with this analysis. D
etailed experimental settings and results for other noise levels, such as 40\%, are provided in the appendix.
\section{Conclusion and Future Work}
This paper presents a new perspective—Decision Pattern Shift (DPS)—for understanding generalization. 
We show that generalization is reflected by the model’s ability to  maintain stable and consistent decision pattern distributions between training and unseen data.
By representing decision patterns and defining the DPS metric, our theoretical and empirical analyses  reveal a strong linear correlation between DPS and the generalization gap across multiple levels. 
Furthermore, analyzing five representative scenarios demonstrates that generalization degradation follows a progressive evolution along the DPS spectrum, offering a unified view of diverse failure modes.

The DPS framework further opens promising avenues for practical applications, including early generalization-risk warning, failure-mode diagnosis, and channel-level defect localization.
In future work, we plan to extend this framework to more modern architectures, such as Vision Transformers, and adapt our method to dense prediction tasks, such as object detection and segmentation, given their inherent reliance on localized classification.

{
    \small
    \bibliographystyle{ieeenat_fullname}
    \bibliography{decision_pattern}
}

\newpage

\appendix

\begin{table*}[t]
  \centering
  \caption{Comparison of intra-class consistency and inter-class confusability across different models and datasets.
  Decision patterns achieve consistently higher intra-class consistency and lower inter-class confusability than activation patterns on almost all datasets and architectures.}
  \vspace{3pt}
  \label{tab:appendix_intra_inter}
  \resizebox{\textwidth}{!}{
  \begin{tabular}{cc|cccc|cccc}
  \toprule
  \multirow{2}{*}{\textbf{Model}} & \multirow{2}{*}{\textbf{Patterns}} & 
  \multicolumn{4}{c|}{Intra-class consistency $\uparrow$} &
  \multicolumn{4}{c}{Inter-class confusability  $\downarrow$} \\
  \cline{3-10}
  & & CIFAR-10 & CIFAR-100 & TinyImageNet & ImageNet & CIFAR-10 & CIFAR-100 & TinyImageNet & ImageNet \\
  \midrule
  \multirow{2}{*}{\textbf{VGG16}} & Activation 
  & 0.976 & 0.970 & 0.586 & 0.722 & 0.422 & 0.612 & \textbf{0.463} & 0.520  \\
  & Decision  & \textbf{0.989}   & \textbf{0.993} & \textbf{0.845} & \textbf{0.835} & \textbf{0.259} & \textbf{0.577} & 0.492 & \textbf{0.415}  \\
  \midrule
  \multirow{2}{*}{\textbf{GooLeNet}} 
  & Activation  & 0.926 & 0.917 & 0.823 & 0.810 & 0.394 & 0.470 & 0.652 & 0.664 \\
  & Decision  &  \textbf{0.956} & \textbf{0.980} & \textbf{0.934} & \textbf{0.929}  & \textbf{0.189} 
  & \textbf{0.268} & \textbf{0.377} & \textbf{0.381} \\
  \bottomrule
  \end{tabular}
  }
  \end{table*}

\section{Appendix}
In this appendix, we provide additional implementation details and experimental results referenced in the main text.

\begin{wrapfigure}{r}{0.58\columnwidth}
  \centering
  \begingroup
  \def\svgwidth{0.48\textwidth}%
\begingroup%
  \makeatletter%
  \providecommand\color[2][]{%
    \errmessage{(Inkscape) Color is used for the text in Inkscape, but the package 'color.sty' is not loaded}%
    \renewcommand\color[2][]{}%
  }%
  \providecommand\transparent[1]{%
    \errmessage{(Inkscape) Transparency is used (non-zero) for the text in Inkscape, but the package 'transparent.sty' is not loaded}%
    \renewcommand\transparent[1]{}%
  }%
  \providecommand\rotatebox[2]{#2}%
  \newcommand*\fsize{\dimexpr\f@size pt\relax}%
  \newcommand*\lineheight[1]{\fontsize{\fsize}{#1\fsize}\selectfont}%
  \ifx\svgwidth\undefined%
    \setlength{\unitlength}{1530.57202148bp}%
    \ifx\svgscale\undefined%
      \relax%
    \else%
      \setlength{\unitlength}{\unitlength * \real{\svgscale}}%
    \fi%
  \else%
    \setlength{\unitlength}{\svgwidth}%
  \fi%
  \global\let\svgwidth\undefined%
  \global\let\svgscale\undefined%
  \makeatother%
  \begin{picture}(1,1.49790991)%
    \lineheight{1}%
    \setlength\tabcolsep{0pt}%
    \put(0,0){\includegraphics[width=\unitlength,page=1]{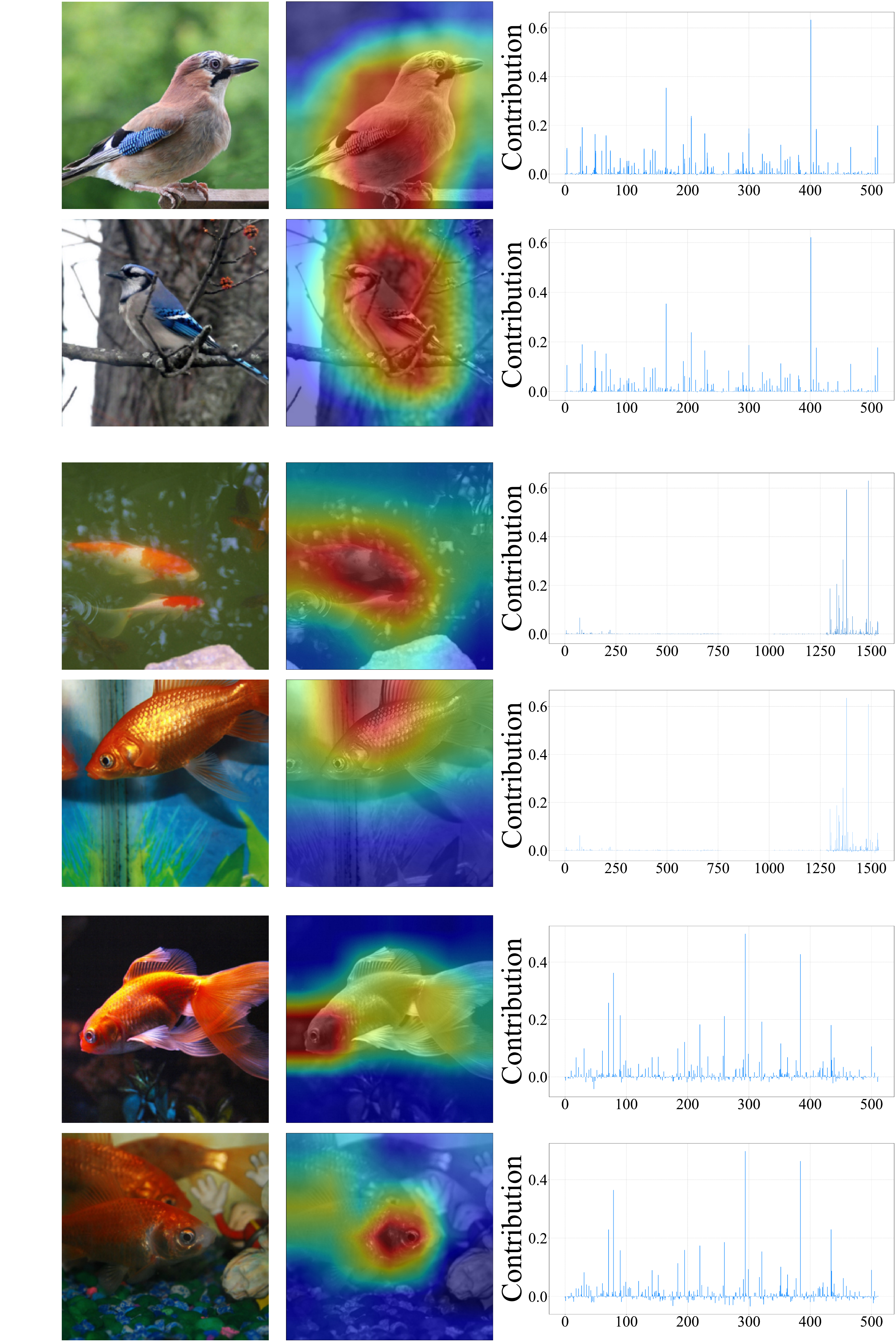}}%
    \put(-0.00166815,1.4670685){\color[rgb]{0,0,0}\makebox(0,0)[lt]{\lineheight{1.25}\smash{\begin{tabular}[t]{l}(a)\end{tabular}}}}%
    \put(-0.00166815,0.93638455){\color[rgb]{0,0,0}\makebox(0,0)[lt]{\lineheight{1.25}\smash{\begin{tabular}[t]{l}(b)\end{tabular}}}}%
    \put(-0.00166815,0.44784171){\color[rgb]{0,0,0}\makebox(0,0)[lt]{\lineheight{1.25}\smash{\begin{tabular}[t]{l}(c)\end{tabular}}}}%
  \end{picture}%
\endgroup%
  \endgroup

 \vspace{-10pt}
    \caption{Visualization of decision patterns across different architectures on ImageNet. 
    For each image, we show: (left) the input image, (middle) its GradCAM heatmap, and (right) the corresponding decision pattern. 
    Results are presented for three architectures: (a) ResNet34, (b) GoogLeNet, and (c) VGG16. }
  \label{fig:appendix_DP_illustration}
  \vspace{-25pt}
\end{wrapfigure}
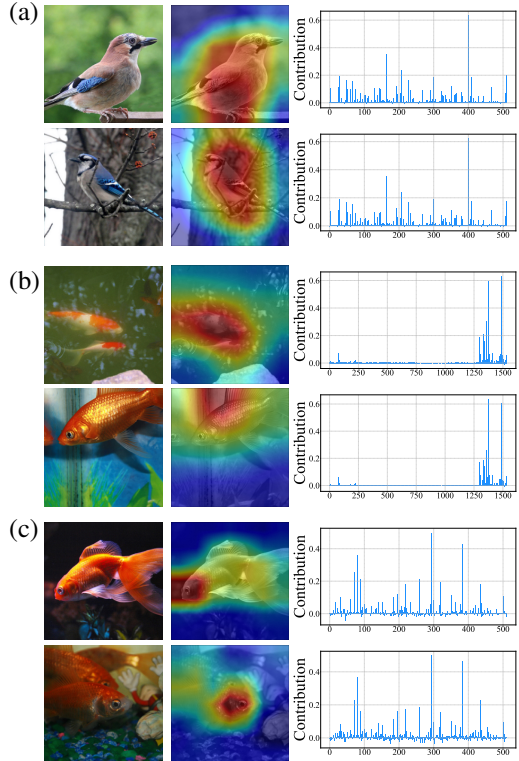

\subsection{Additional details of formulation}
\label{sec:Additional details of formulation}
Since the decision pattern vectors (i.e., channel-wise contribution) of different samples may have varying magnitudes, directly comparing them can be misleading. 
To ensure comparability across samples, we apply an $\ell_2$ normalization to each decision pattern vector:
\begin{equation}
    \boldsymbol{q}^c(x) = \frac{\boldsymbol{p}^c(x)}{\Vert \boldsymbol{p}^c(x) \Vert}
\end{equation}
After normalization, the class-level average decision pattern for class $c$ is  defined as:
\begin{equation}
    \boldsymbol{\overline p}^c = \mathbb{E}_{x \in \mathcal{D}^c_{train}} \left[\boldsymbol{q}^c(x)\right]
\end{equation}
This normalization ensures that the resulting decision patterns reflect the differences among samples without being confounded by variations of their overall magnitude.

\subsection{Additional visualization results of decision patterns}
\label{sec:Additional visualization results of decision patterns}
In the main text, we presented the 64-dimensional decision patterns of two dog images from CIFAR10 using ResNet20.
Here, we extend the visualization to ImageNet and include additional architectures: ResNet34, VGG16 (512 dimensions), and GoogLeNet (1536 dimensions).

As shown in Fig.~\ref{fig:appendix_DP_illustration}, we observe consistent phenomena across all models.
Only a small subset of channels exhibits strong contributions, while most remain weak or negligible.
The sparsity is especially notable in GoogLeNet, where only a few channels dominate the overall contribution.

Furthermore, even in high-dimensional spaces, decision patterns of same-class samples still exhibit highly similar structures.
This indicates that images with significantly different visual appearances are mapped to highly consistent and low-dimensional decision pattern within the network.

\subsection{Additional Results on intra-class consistency and inter-class separability}
\label{sec:Additional Results on intra-class consistency and inter-class separability}
We evaluate the intra-class consistency and inter-class separability of decision patterns across four datasets: CIFAR10, CIFAR100 \cite{krizhevsky2009learning}, TinyImageNet \cite{Le2015TinyIV}, and ImageNet \cite{deng2009imagenet}. The evaluation covers several representative architectures, including VGG16 \cite{simonyan2014very}, ResNet \cite{he2016deep}, and GoogLeNet \cite{szegedy2015going}.
While the main text reports only a subset of the quantitative results, 
we provide the full set of results and more visualizations here to offer a more comprehensive view. 

\begin{figure*}[t]
  \centering
  \begingroup
  \def\svgwidth{0.925\textwidth}%
  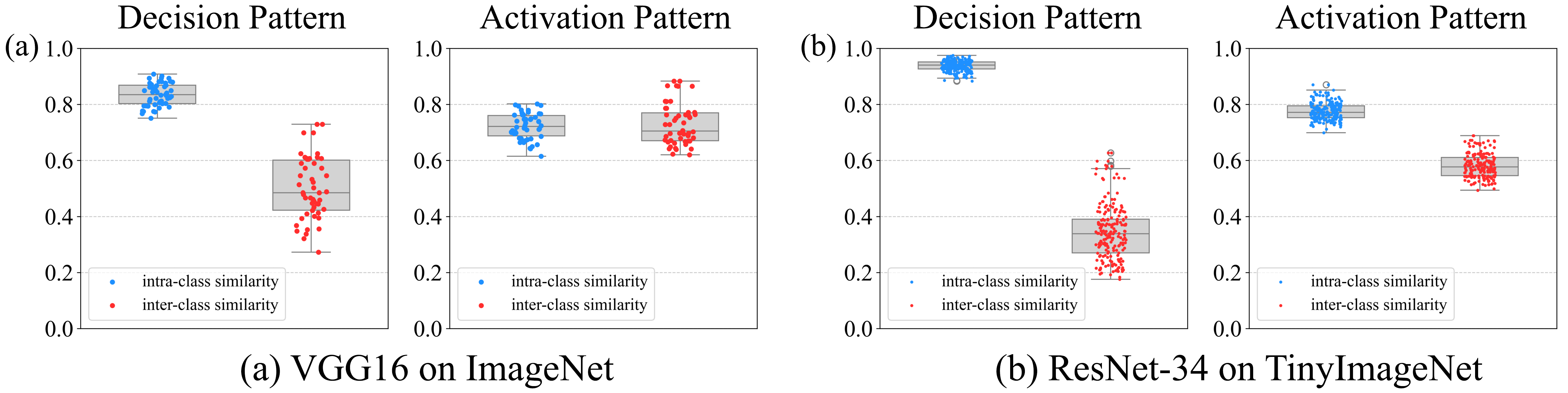%
  \endgroup

     \caption{Box-plots of intra-class and inter-class similarities for decision patterns and activation patterns across different models: (a) VGG-16 on ImageNet. (b) ResNet34 on TinyImageNet. Each point corresponds to one class.}
  \label{fig:appendix_intra_inter}
\end{figure*}

As shown in Tab.~\ref{tab:appendix_intra_inter}, our proposed decision pattern achieves consistently 
high intra-class consistency across nearly all datasets and architectures, with average cosine similarity typically exceeding 0.9 and the lowest value still above 0.83. 
At the same time, it exhibits very low inter-class confusability—almost always below 0.5—substantially outperforming activation patterns.
These differences are further visualized as box plots (see Fig.~\ref{fig:appendix_intra_inter}), clearly revealing the superior consistency  and class discriminability achieved by decision patterns.

\begin{figure*}[t]
  \centering
  \begingroup
  \def\svgwidth{0.925\textwidth}%
  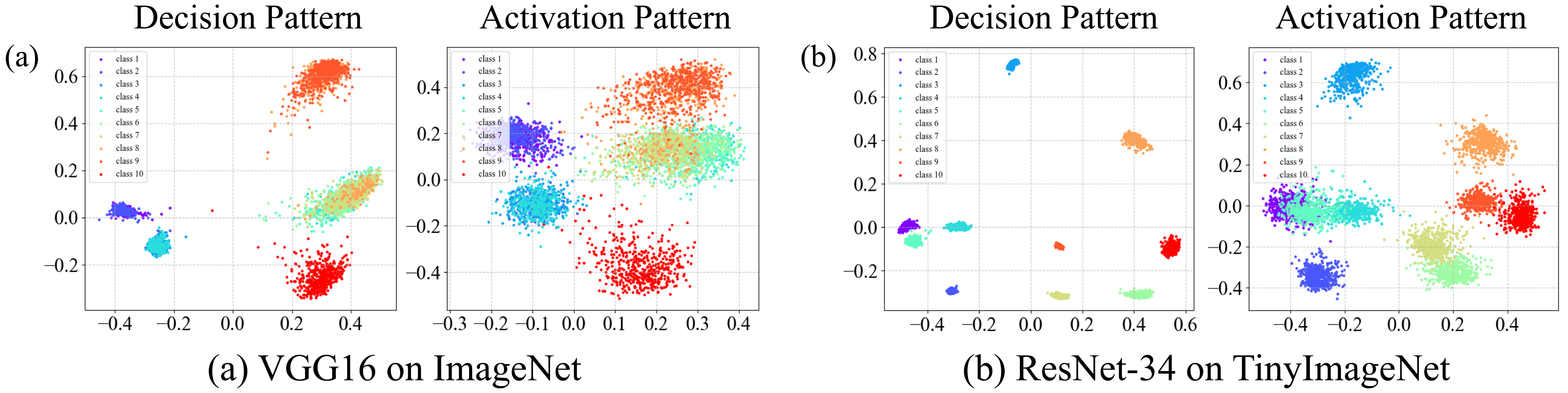%
  \endgroup

     \caption{PCA projections of decision pattern vectors and activation pattern vectors across two models: (a) VGG16 on ImageNet. (b) ResNet-34 on TinyImageNet. Each color denotes one class.
     Decision patterns form tightly clustered and well-separated class structures, whereas activation patterns exhibit markedly greater intra-class dispersion and stronger inter-class overlap. }
  \label{fig:appendix_intra_inter_pca}
\end{figure*}

We further provide additional visualizations to examine the underlying structure of the decision-pattern space.
Fig.~\ref{fig:appendix_intra_inter_pca} presents PCA projections of 10 classes from ImageNet  (left) and  TinyImageNet (right).
Decision patterns form highly compact and well-separated clusters, where each cluster  exactly corresponds to a class.
In contrast, activation patterns yield much more dispersed intra-class distributions and exhibit more frequent overlaps across classes.

These results demonstrate a consistent conclusion:
\textbf{decision patterns form a highly cohesive, well-separated, and semantically aligned decision space, whereas activation patterns do not.}

\subsection{Additional results on limitations}
\label{sec:Additional results on limitations}

To empirically validate the limitations of Decision Pattern Shift (DPS) under label noise discussed in the main text, we train VGG16 from scratch on CIFAR100 with varying proportions of randomly corrupted training labels (0\%, 20\% and 40\%). 
As established, DPS relies on the class-average decision pattern $\overline{\boldsymbol{p}}^{\,c}$ computed from training samples as a reference. However, under significant label noise, the model fails to learn a stable and correct average decision pattern. 

\begin{figure}[h]
  \centering
  \includegraphics[width=\textwidth]{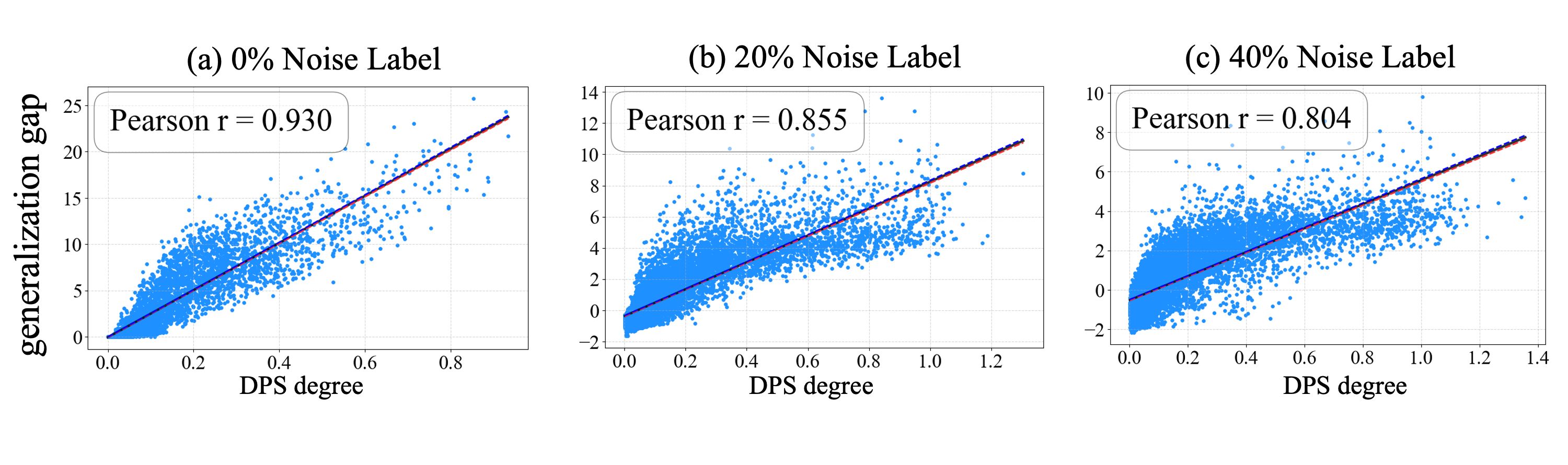}
  \vspace{-30pt}
  \caption{Correlation between Decision Pattern Shift (DPS) and generalization gap on CIFAR100 with VGG16 under varying levels of label noise (0\%, 20\% and 40\%). 
  As the proportion of noisy labels increases, the class-average reference pattern becomes less reliable, leading to a visible degradation in the linear correlation.}
  \label{fig:appendix_noise}
\end{figure}

Fig.~\ref{fig:appendix_noise} visualizes how this degradation affects the correlation between DPS and the generalization gap. As the noise ratio increases, the previously strong linear correlation visibly drops: from 0.930 (0\% noise) to 0.855 (20\% noise), and continues to decrease at 40\% noise levels. These results confirm that the reliability of DPS is heavily dependent on the model's ability to form coherent class-level decision logic, and that deviations from a corrupted reference do not necessarily indicate true generalization failure.

\subsection{Comparison of activation pattern and decision pattern on linear correlation}
\label{sec:Comparison of activation pattern and decision pattern on linear correlation}
\begin{figure}[h]
  \centering
 \includegraphics[width=0.8\textwidth]{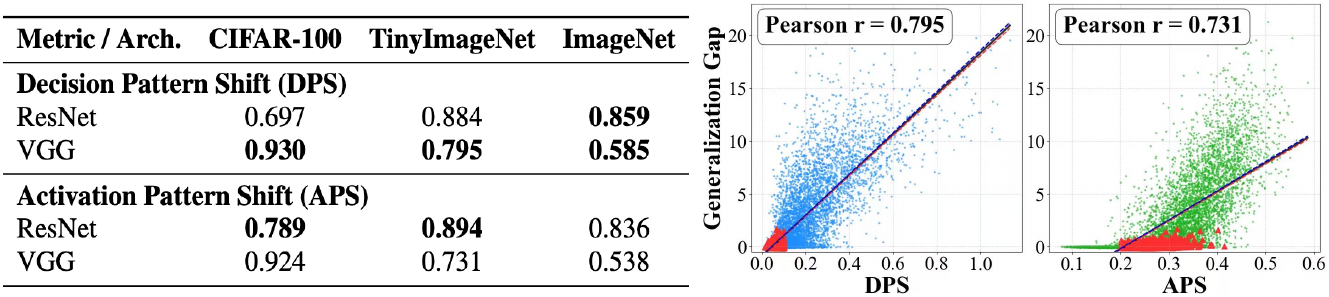}
 \vspace{-5pt}
 \caption{Linear correlation with the generalization gap: comparison between sample-level activation patterns and decision patterns (left), and representative results on TinyImageNet using VGG16 (right).}
  \label{fig:DPS_vs_APS}
\end{figure}

As shown in Fig.~\ref{fig:DPS_vs_APS}, the activation pattern shift (APS) and decision pattern shift (DPS) exhibit significantly different behaviors with respect to generalization, despite having comparable overall correlations with generalization error.
Specifically, APS reflects benign representation drift across datasets and architectures, manifested as many correctly classified test samples with substantial activation shifts but small decision pattern shifts (red triangles); in Fig.~\ref{fig:DPS_vs_APS}, these account for nearly 40\% of correct predictions.
In contrast, DPS suppresses such benign variations and more selectively highlights decision-relevant drift.
Moreover, activation pattern is not faithful to the output prediction of the model mathematically.
This observation indicates that \textbf{activation (in)stability alone is insufficient} to characterize generalization behavior, while decision (in)stability provides \textbf{complementary, decision-centric} information.

\end{document}